%% file: main.tex
\documentclass{article}
\usepackage{iclr2027_preprint,times}
\usepackage{amsmath,amssymb,booktabs,graphicx,multirow}
\usepackage{listings}
\usepackage{colortbl}
\definecolor{ordergray}{gray}{0.94}
\usepackage{algorithm}
\usepackage{float}
\usepackage{placeins}
\usepackage{algpseudocode}
\usepackage{wrapfig}
\usepackage{ragged2e}
\usepackage{microtype}
\usepackage{hyperref}
\usepackage{url}
\hypersetup{hidelinks}
\graphicspath{{figures/}}

\title{When Skills Remain but Access Fails:\\
Factorized Skill Memory for Continual Skill Internalization}
\author{%
{\fontsize{9.2}{11}\selectfont\bfseries Haomiao Qiu$^{1,2,\ast}$ \hspace{0.4em} Yining Tang$^{3,\ast}$ \hspace{0.4em} Hongyi Wan$^{1}$ \hspace{0.4em} Changfeng Sun$^{1}$ \hspace{0.4em} Ziyue Qiao$^{2,\dagger}$ \hspace{0.4em} Miao Zhang$^{1,\dagger}$}\\[0.5em]
{\fontsize{8.0}{9.6}\selectfont\normalfont $^{1}$ Harbin Institute of Technology, Shenzhen \hspace{0.8em} $^{2}$ Great Bay University \hspace{0.8em} $^{3}$ Zhejiang University}\\[0.25em]
{\fontsize{8.3}{10}\texttt{24B951058@stu.hit.edu.cn}}%
}

\begin{document}
\maketitle

\begingroup
\renewcommand{\thefootnote}{\fnsymbol{footnote}}
\footnotetext[1]{Equal contribution;\quad $^{\dagger}$ Co-corresponding authors.}
\endgroup

\begin{abstract}
Skill internalization aims to convert externally specified procedural skills into persistent model capabilities, allowing language-model agents to execute them without retrieving or reinjecting skill descriptions at inference time. Yet when new skills are learned continually, whether previously internalized skills remain usable is still poorly understood. We study forgetting in continual skill internalization. Using correct-skill-address reactivation as a diagnostic, we identify a hierarchical pattern: closed-book execution can collapse while supplying the correct skill address substantially restores learned behavior, and continued optimization later degrades execution even under the correct skill address. We call these two regimes \emph{skill-addressing forgetting} and \emph{skill-execution forgetting}. Guided by this finding, we introduce \textbf{Factorized Skill Memory (FSM)}, which separates continual skill memory into a plastic \emph{Address Memory} and a stable \emph{Execution Memory}. At each stage, FSM trains a new LoRA executor using only current-stage data, freezes it after acquisition, and appends it to Execution Memory; in parallel, Address Memory incrementally learns to map current requests to executors without replaying historical requests or execution trajectories. At inference time, Address Memory selects one executor from the raw request, and only that frozen executor performs a single language-model generation. On a five-stage benchmark built from 100 TaskBench skills, FSM achieves 65.95\% mean exact match and 93.38\% shard-routing accuracy across two task orders, improving over Progressive-LoRA by 14.3 percentage points while remaining within 1.9 points of Replay, which retains historical training examples. These results support an Address--Execution view of continual skill retention, in which stable procedural content and reliable access are distinct design problems.
\end{abstract}

\input{sections/01_introduction}
\input{sections/02_related_work}
\input{sections/03_benchmark_protocol}
\input{sections/04_reactivation_phenomenon}
\input{sections/05_factorized_skill_memory}
\input{sections/06_main_results}
\input{sections/10_conclusion}

\paragraph{AI Use Statement.}

AI-assisted tools were used for language polishing, data cleaning and consistency checks, and figure preparation and refinement. All scientific claims, experimental results, analyses, and final manuscript content were reviewed and verified by the authors.

\paragraph{Reproducibility Statement.}
The benchmark protocol, prompts, metrics, training and inference settings,
FSM procedure, and supporting controls are documented in the manuscript and
appendix.

\bibliography{references}
\bibliographystyle{iclr2027_conference}

\clearpage
\appendix
\setcounter{figure}{0}
\setcounter{table}{0}
\renewcommand{\thefigure}{A\arabic{figure}}
\renewcommand{\thetable}{A\arabic{table}}
\renewcommand{\theHfigure}{A\arabic{figure}}
\renewcommand{\theHtable}{A\arabic{table}}

\input{sections/appendix_a_reproducibility}

\FloatBarrier
\input{sections/appendix_d_additional_audits}
\FloatBarrier
\input{sections/08_analysis_limitations}

\end{document}

%% file: sections/01_introduction.tex
\section{Introduction}
Large language model agents increasingly rely on reusable \textbf{Skills} to
organize tool use, environment interaction, and multi-step action generation
\citep{yao2023react,schick2023toolformer,shen2024taskbench}. A Skill packages
procedural guidance into a workflow: which operations to invoke, how their
inputs are constructed, and how intermediate steps depend on one another
\citep{wang2023voyager,yang2026survey}. Most skill-based systems still keep
this guidance outside the model and retrieve the relevant Skill text at
inference time. Execution therefore depends both on retrieval and on the
model's ability to follow the retrieved description. Long contexts are used unevenly, irrelevant context can distract reasoning,
and detailed Skill documentation may not reliably improve execution
\citep{liu2024lostmiddle,shi2023distracted,li2026skillsbench}. Recent work on
\textbf{Skill Internalization} moves this procedural guidance into model
parameters, allowing learned behavior to be used without repeatedly injecting
the Skill description
\citep{lu2026skill0,lin2026skillc,zhang2026skilltolora,
yu2026latentskill}.

We consider agents that acquire new Skills over time. In this
continual-learning setting, disjoint Skill families arrive sequentially, the
model internalizes these Skills in sequence, and the final model receives only
the user request at inference time, without a Skill family label, Skill name,
or retrieved Skill card, while retaining the capabilities of all earlier
Skills
\citep{parisi2019continual,delange2022continual,vandeven2019scenarios}. Existing skill-internalization studies do not establish whether earlier Skills remain usable after later Skills are learned. During continual Skill Internalization training, old-Skill closed-book performance can fall to zero within only a few optimization steps after training moves to a new Skill family. Successful internalization does not by itself ensure retention.

The collapse in closed-book performance hides a more structured failure. At a checkpoint that fails all 408 old-task requests, supplying the \textbf{correct skill address}, given by the benchmark's exact procedure identifier or identifiers required by the request, restores execution performance from 0 to 69.85\%. Because this address also reveals the target node identity, we use it as a diagnostic rather than an information-neutral access probe. With further optimization, execution under the correct skill address degrades, and the same ordering appears across SFT, GRPO, and SKILL0. We call the two regimes skill-addressing forgetting and skill-execution forgetting: closed-book execution fails first while execution under the correct skill address remains recoverable, before that recoverability itself erodes.

\begin{figure*}[!t]
  \centering
  \includegraphics[width=\textwidth]{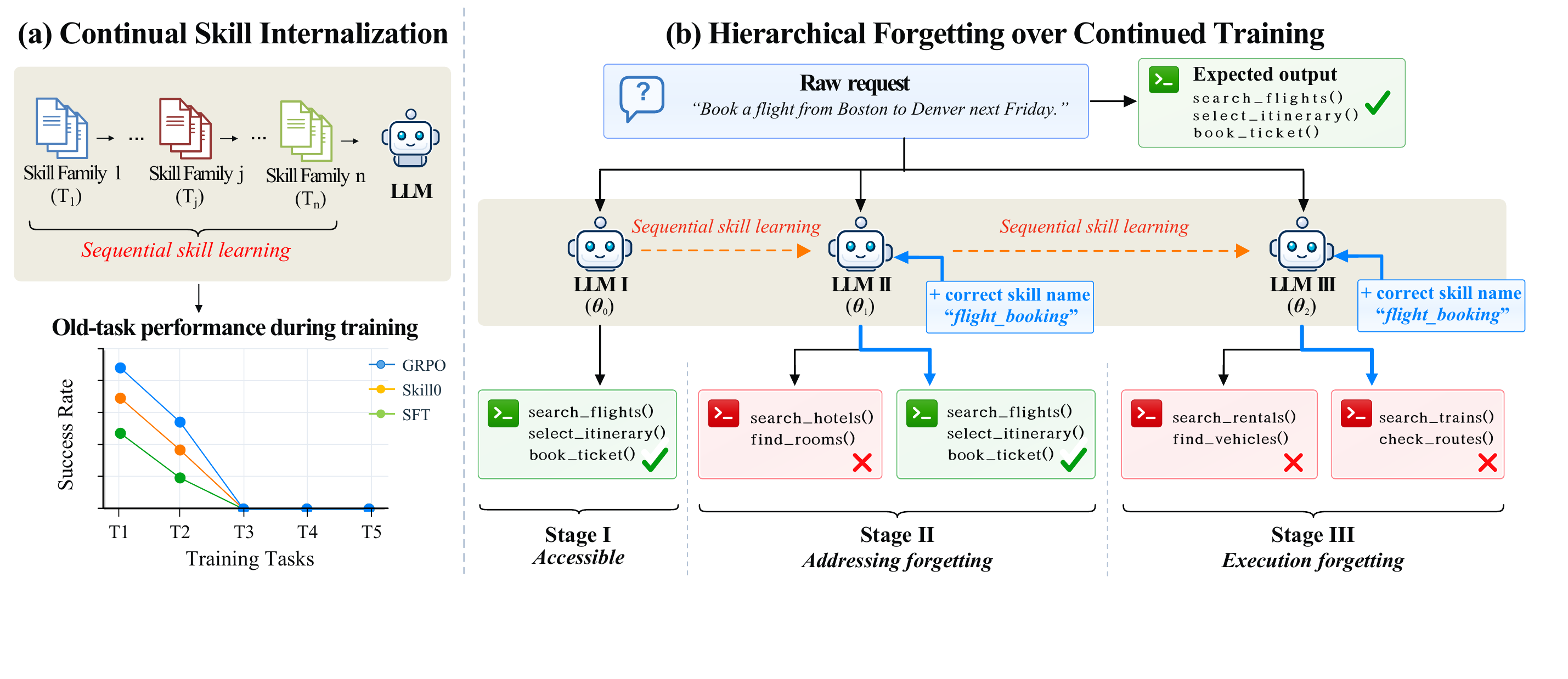}
  \vspace{-30pt}
  \caption{Continual skill internalization and hierarchical forgetting.
  \textbf{(a)} Sequential learning can erase old-task closed-book performance
  within a few optimization steps. \textbf{(b)} A matched request-level example
  exposes two stages of forgetting. During addressing forgetting, the correct
  Skill name recovers execution that the raw request no longer reaches.
  Continued optimization can later damage execution under the correct skill address
  itself.}
  \label{fig:hierarchical-forgetting-overview}
  \vspace{-8pt}
\end{figure*}

Guided by this finding, we introduce \textbf{Factorized Skill Memory (FSM)}, which separates continual skill memory into a plastic \textbf{Address Memory} and a stable \textbf{Execution Memory}. The reactivation diagnostic exposes the Address--Execution split at individual-skill granularity; FSM operationalizes the same split more coarsely, using stage-level shards that each contain multiple Skills. At each stage, FSM trains and freezes one new LoRA executor while Address Memory learns request-to-shard associations without retaining historical requests or execution trajectories. The selected executor still identifies and composes the required Skills within its shard. At inference, Address Memory selects one shard from the raw request, and only that frozen executor performs a language-model generation.

We make three contributions:
\begin{itemize}
\setlength{\itemsep}{1pt}
\setlength{\parskip}{0pt}
\setlength{\parsep}{0pt}
\setlength{\topsep}{2pt}
\item We construct a five-stage benchmark for \textbf{continual skill
internalization} by partitioning 100 TaskBench Skills into disjoint Skill
families and evaluating sequential acquisition, retention, and autonomous
closed-book access.
\item We identify a \textbf{hierarchical forgetting} pattern under
canonical-name reactivation: closed-book execution can collapse while
execution under the correct skill address remains recoverable, before that recoverability
also erodes.
\item We introduce \textbf{FSM}, which separates plastic Address Memory from
frozen Execution Memory. Across two task orders, FSM reaches 65.95\% mean
performance and 93.38\% shard-routing accuracy, improving over the strongest
replay-free parameter-isolation baseline by 14.3 points while remaining within
1.9 points of Replay.
\end{itemize}

%% file: sections/02_related_work.tex
\section{Related Work}

\paragraph{Skill Internalization.}
Agent Skills are commonly represented as procedural artifacts that are
retrieved and inserted into the model context at inference time
\citep{li2026skillsbench,jiang2026demystifying}. Recent work studies
how to convert this guidance into persistent model behavior. SKILL0 and SkillC
learn to execute Skills with progressively less dependence on explicit Skill
context \citep{lu2026skill0,lin2026skillc}. Another line of work parameterizes procedural knowledge directly: Skill-to-LoRA and ParametricSkills
map Skills into adapter parameters, while LatentSkill and PEAM move textual or
experiential Skill information into in-weight representations
\citep{zhang2026skilltolora,zhao2026parametricskills,yu2026latentskill,
guo2026peam}. LifeSkill further studies online lifelong adaptation from
Skill-guided experience \citep{mao2026lifeskill}. Together, these methods show
that procedural guidance can be transferred from context into model
parameters, and several introduce mechanisms for accumulating such
capabilities over time. Our focus is the failure structure that emerges when
internalized Skills are learned sequentially without Skill identity at test
time. We distinguish loss of request-side access from loss of execution
itself, and use this Address--Execution hierarchy to determine what
should remain plastic and what should be protected during continual
internalization.

\paragraph{Continual Learning.}
Continual learning studies how models acquire new tasks without overwriting
earlier competence. Replay preserves old examples, regularization constrains
parameter changes, and architectural methods isolate or expand parameters for
new tasks
\citep{kirkpatrick2017overcoming,li2017learning,lopezpaz2017gradient,
rebuffi2017icarl,rusu2016progressive}. The last family is especially relevant
when different tasks require distinct behaviors. Adapter-based continual
learning allocates task-specific parameters
\citep{madotto2021continual}; EPI combines private parameter-efficient modules
with a non-parametric task identifier so that the appropriate module can be
recovered without storing old training data \citep{wang2023epi}; and MoCL
continually adds and composes modules to preserve knowledge while
supporting transfer \citep{wang2024mocl}. FSM falls within this modular parameter-isolation line, with its decomposition
motivated by the failure structure of continual Skill internalization. In continual Skill internalization, our reactivation
experiments show that request-to-Skill access can fail while conditional
execution remains intact. FSM therefore isolates the acquired procedural
execution in frozen multi-Skill shards and treats access as the
updated memory state. 

%% file: sections/03_benchmark_protocol.tex
\section{Continual Skill Internalization}
\label{sec:benchmark}

\paragraph{Problem setting.}
We consider a sequence of stages $t=1,\ldots,T$, where each stage introduces a
previously unseen skill set $\mathcal S_t$ and corresponding request--workflow
supervision. The skill sets are pairwise disjoint, and after stage $t$ the
desired competence includes all skills in $\mathcal S_1\cup\cdots\cup\mathcal
S_t$. For a request $x$, successful closed-book execution requires the model to
produce the correct request-conditioned workflow using only the raw request.
Training may use stage-local workflow supervision and, for skill-guided
baselines, textual skill cards. Evaluation is autonomous and closed-book: the
model receives only $x$, not a stage label, skill name, or catalog. This setting tests \emph{acquisition} of new skills and \emph{retention} of
earlier skills under autonomous inference.

\noindent
\begin{minipage}[t]{0.57\linewidth}
\vspace{0pt}
\paragraph{Benchmark.}
We instantiate this stream using the public Microsoft TaskBench benchmark
\citep{shen2024taskbench}, partitioning its 100 canonical skills into five
disjoint semantic families (Table~\ref{tab:benchmark-statistics}). We use each
family as a unit of continual acquisition: a capability package containing
several related procedures. The five packages contain 40, 11, 21, 9, and 19
canonical skills. Requests can require multi-node workflows composed from
several skills in the same package. Each stage contains 1,000 training
requests, for 5,000 in total, and the five held-out sets contain 2,186 requests.
The complete skill inventory and output contract appear in
Appendix~\ref{app:benchmark-metrics}.
\end{minipage}
\hfill
\begin{minipage}[t]{0.40\linewidth}
\vspace{0pt}
\centering
\refstepcounter{table}\label{tab:benchmark-statistics}
{\small Table~\thetable: TaskBench benchmark statistics. Each stage uses 1,000
training requests.}
\par\vspace{2pt}
\fontsize{6.0}{6.6}\selectfont
\setlength{\tabcolsep}{1.8pt}
\renewcommand{\arraystretch}{0.92}
\resizebox{\linewidth}{!}{%
\begin{tabular}{llrr}
\toprule
Task & Semantic family & \# Skills & Eval \\
\midrule
T1 & Daily Life & 40 & 864 \\
T2 & Language / Speech & 11 & 408 \\
T3 & Retrieval / Understanding & 21 & 332 \\
T4 & Vision / Document & 9 & 299 \\
T5 & Generation / Editing & 19 & 283 \\
\midrule
Total & -- & 100 & 2,186 \\
\bottomrule
\end{tabular}}
\end{minipage}

\paragraph{Metrics.}
Our primary metric is \textbf{exact-match task success (EM)}, which requires
the complete prescribed workflow output to be correct; for TaskBench this is
the benchmark's canonical graph exact match, but we refer to it simply as EM in
the main text. The T1--T5 columns and their task-macro average all report this
same quantity. We use three secondary diagnostics to interpret failures:
\textbf{Tool-choice accuracy} checks whether the ordered tool sequence is
correct while ignoring request-specific values; \textbf{Argument-value F1}
compares predicted and target $(\text{skill},\text{argument},\text{value})$
tuples; and \textbf{BWT} measures final old-task EM minus EM immediately after
that task was learned, with zero indicating no forgetting; additional
benchmark-specific metrics appear in the appendix.

\paragraph{Direct continual optimization rapidly eliminates old-task closed-book performance.}
Before introducing any mechanism for preservation, we first measure how direct
continual optimization behaves when training moves away from an acquired task.
Immediately after learning the old task, EM is 77.45\% for SFT, 84.31\% for
GRPO \citep{shao2024deepseekmath}, and 84.07\% for the SKILL0
\citep{lu2026skill0} curriculum. After
optimization switches to the next task, old-task EM falls to zero at observed
steps 8, 6, and 10, respectively, and remains there over the rest of the
measured interval (Figure~\ref{fig:two-stage-forgetting}). Across hundreds of
held-out requests, none of the old-task outputs remain exactly correct. Under the standard closed-book
criterion, all three procedures therefore exhibit catastrophic forgetting of
the previously internalized skill. This endpoint shows that retention has failed without revealing which part of
the acquired capability was lost. Section~\ref{sec:hierarchical-forgetting}
tests this distinction directly.

%% file: sections/04_reactivation_phenomenon.tex
\section{When Forgotten Skills Come Back}
\label{sec:hierarchical-forgetting}

\subsection{Correct-Skill-Address Reactivation}
Section~\ref{sec:benchmark} showed that old-task closed-book performance can
fall to zero after continual optimization moves to a new task. That endpoint
does not reveal whether the old procedure has been erased or whether the model
has lost access to behavior that is still present. We distinguish these cases
with a controlled reactivation intervention. We take a post-forgetting
checkpoint that scores zero on all 408 closed-book T2 requests and supply only
the \emph{correct skill address}, given by the benchmark's exact procedure identifier or identifiers required by each request. On the same frozen model,
\textbf{full-task EM recovers from 0 to 69.85\%.} The address provides no
argument values, workflow steps, or target serialization, so the model must
still construct the request-conditioned workflow. Because it also reveals the
gold node identity, we use the correct skill address as a diagnostic rather
than a pure access probe. Additional controls are reported in
Appendix~\ref{app:skill-name-controls}.

A frozen-base control further separates address recognition from learned
execution. On 408 T2 requests, frozen Qwen3-8B \citep{yang2025qwen3} selects
the correct skill on 92.65\% of examples when given the correct skill address
but obtains 0\% full-task EM; by contrast, the post-forgetting checkpoint
recovers 69.85\% under the same address after request-only performance reaches zero. Recognizing the supplied address alone therefore cannot account
for the recovered behavior.
We use \emph{Address} for access to the relevant procedural state and
\emph{Execution} for producing the request-conditioned workflow once that
state is available. This reversal shows that substantial execution
remains available once the correct procedure identity is supplied; we call
this regime \emph{skill-addressing forgetting}.

\subsection{Execution under the correct skill address erodes later}
\begin{figure*}[!t]
  \centering
  \includegraphics[width=\textwidth]{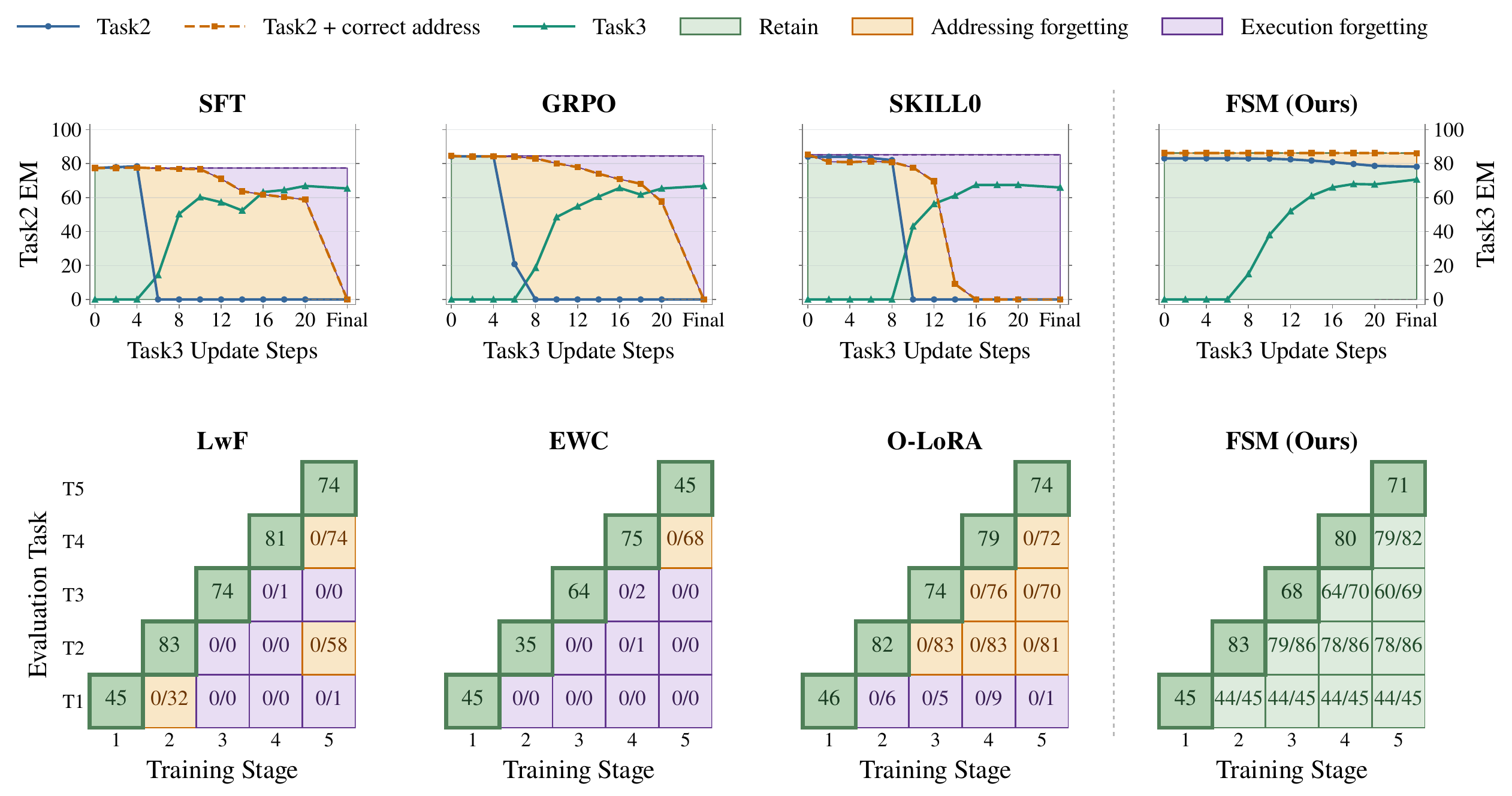}
  \caption{Hierarchical forgetting during continual skill internalization.
  \textbf{Top:} during the T2$\rightarrow$T3 transition, closed-book Task2 EM
  can collapse while Task2 remains recoverable when the correct skill address
  is supplied; with continued optimization, execution under the correct skill address
  can also erode. The frozen-executor diagnostic preserves the old task while
  learning Task3. \textbf{Bottom:} five-stage evaluation matrices for four
  continual-learning conditions. Off-diagonal cells report closed-book /
  + Correct Skill Address EM. Across both rows, pale green denotes retained performance,
  pale yellow denotes addressing forgetting, and pale purple denotes execution
  forgetting. Diagonal cells show task-acquisition EM. The FSM column is our
  method.}
  \label{fig:two-stage-forgetting}
\end{figure*}

To see what happens after this access failure,
Figure~\ref{fig:two-stage-forgetting} follows the first 20 optimizer steps of
the T2$\rightarrow$T3 transition, evaluating each checkpoint closed-book and
with the correct skill address. When closed-book T2 first reaches zero,
full-task EM under the correct skill address is still 76.96\% for SFT, 84.07\% for GRPO, and
77.70\% for SKILL0. Because the paired measurements come from the same checkpoint, this gap shows
a loss of autonomous closed-book access relative to the
target-identity-assisted condition while conditional execution remains
available. Further T3 optimization then reduces performance under the correct skill address from 77.45\% to 58.82\% for SFT, from 84.56\% to 57.60\% for
GRPO, and from 85.29\% to zero for SKILL0. We call this later failure
\emph{skill-execution forgetting}: even the correct address can no longer
recover the earlier behavior.

The bottom matrices show the same ordering across other continual-learning
methods: closed-book performance can collapse before execution under the
correct skill address. This diagnostic operates at individual-skill
granularity. FSM applies the same functional separation at shard granularity:
routing selects a multi-skill capability package, while the selected executor
still performs within-shard skill selection, composition, and execution.

%% file: sections/05_factorized_skill_memory.tex
\section{Factorized Skill Memory}
\label{sec:fsm}

Section~\ref{sec:hierarchical-forgetting} separates request-side access from
execution at individual-skill granularity. FSM preserves this functional split
without allocating one executor per skill: related skills are grouped into
stage-level execution shards, and Address Memory learns only which shard to
read. Formally,
\[
\mathcal M_t=(A_t,\mathcal E_t),
\]
where $A_t$ is \textbf{Address Memory} and
$\mathcal E_t=\{E_1,\ldots,E_t\}$ is \textbf{Execution Memory}. Address remains
plastic as new skill families arrive, whereas Execution Memory is append-only.
At stage $t$, FSM trains and freezes a fresh executor $E_t$ on $\mathcal S_t$;
current requests update Address with shard identity $t$ and are then discarded.
Historical requests and workflow trajectories are never replayed.

\begin{figure*}[!t]
  \centering
  \vspace{-6pt}
  \includegraphics[width=\textwidth]{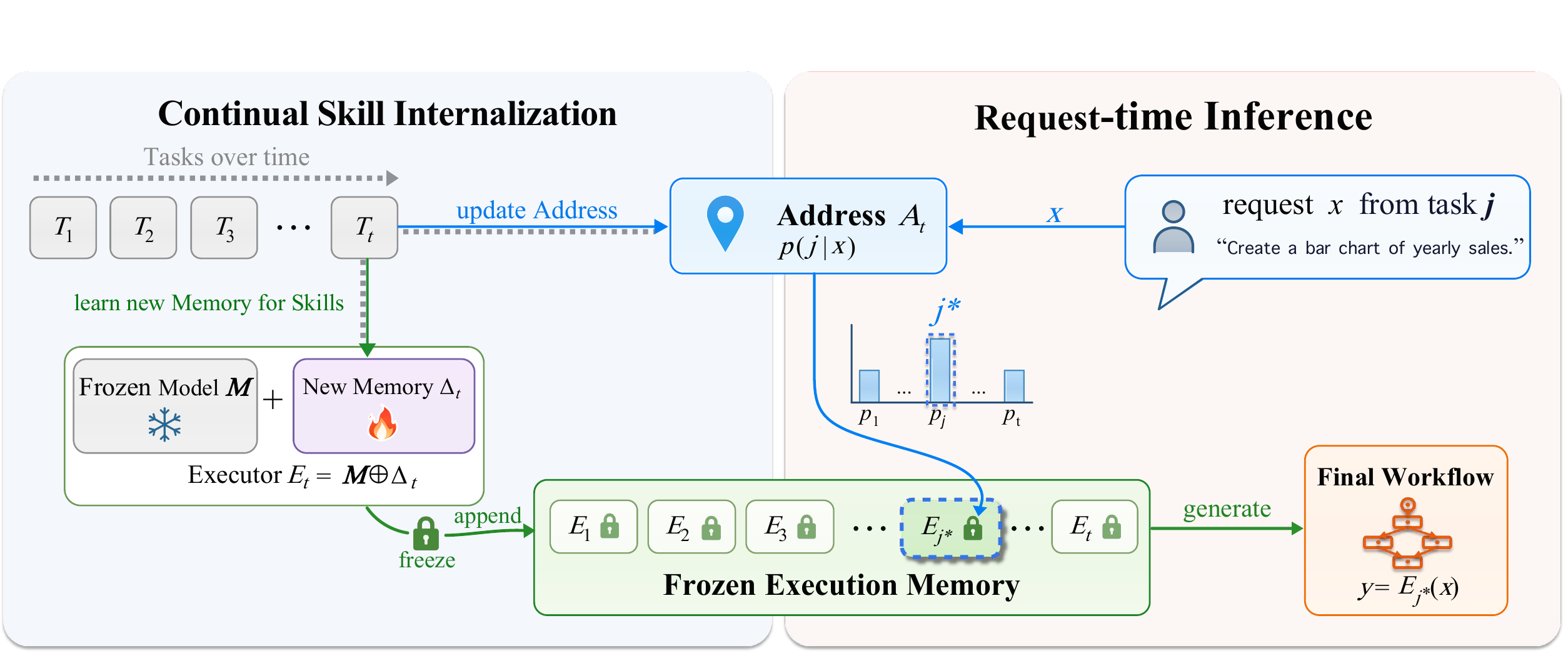}
  \caption{Factorized Skill Memory (FSM). \textbf{Left: memory write.}
  At stage $t$, current requests update Address Memory while a fresh executor
  $E_t=\mathcal B\oplus\Delta_t$ is trained from the frozen base and a
  stage-specific LoRA. The completed executor is frozen and appended to
  Execution Memory as a new skill-family shard. \textbf{Right: memory read.}
  Address maps the raw request to $p(j\mid x)$, selects $j^\star$, and reads
  one frozen execution shard. Only $E_{j^\star}$ performs language-model
  generation.}
  \label{fig:mechanism}
  \vspace{-8pt}
\end{figure*}

Each shard contains 9--40 canonical skills, so shard routing only narrows the
execution space to a capability package. It does not replace within-shard skill
selection: the chosen executor must still recover the required skill sequence,
request-specific arguments, and dependencies from the raw request. At
inference, Address selects one shard and only that executor performs
language-model generation. Figure~\ref{fig:mechanism} summarizes the
write-and-read process; Appendix~\ref{app:benchmark-metrics} provides further details.

\subsection{Execution Memory: Immutable Parametric Shards}

Let $B$ denote the shared frozen base model. At stage $t$, FSM initializes a
fresh LoRA executor $E_t=B\oplus\Delta_t$, trains $\Delta_t$ only on
current-stage skill-execution supervision, freezes it after acquisition, and
appends it to the bank, $\mathcal E_t=\mathcal E_{t-1}\cup\{E_t\}$. No
subsequent stage updates an earlier $\Delta_j$. This boundary follows the
failure pattern in Section~\ref{sec:hierarchical-forgetting}: conditional
execution can survive after request-side access has failed, so FSM protects the
parameters that store execution and places subsequent adaptation in Address
Memory. Low-rank adapters provide a compact implementation of these isolated
parametric writes \citep{hu2022lora,rebuffi2017learning,houlsby2019adapters}.

Each shard stores a family of related skills rather than a single procedure.
For example, the T1 shard contains 40 canonical skills; selecting T1 only
determines which execution memory to read, while the executor still chooses and
composes the required skills and produces the complete request-conditioned
graph. The shared 8.2B backbone is stored once. Under our rank-16 LoRA
configuration on the $q/k/v/o$ projections, each Qwen3-8B shard adds 15.34M
parameters, about 0.19\% of the backbone, and the complete five-shard bank adds
76.68M parameters, about 0.94\%. The five BF16 LoRA deltas occupy about 146 MiB, and inference activates only
the selected LoRA.

\subsection{Address Memory: Continual Associative Indexing}

With earlier execution shards fixed, the continually updated state is the
request-to-shard index
\[
p_{\mathrm{Address}}(j\mid x),\qquad j\in\{1,\ldots,t\}.
\]
The test request contains no task label, correct skill address, or skill
catalog, so shard identity must be recovered from request text alone. Our
deployed Address maps request text into a fixed-dimensional hashing
representation and updates an incremental Naive Bayes classifier using only the
current request pool with label $t$. After the update, those requests are
discarded; the retained state consists only of class and feature statistics.
Address Memory can therefore absorb new request-to-shard associations without
retaining historical requests while the executor bank stays fixed.

\subsection{Single-Shard Memory Read}

At inference, FSM performs a hard memory read:
\[
j^\star=\arg\max_j p_{\mathrm{Address}}(j\mid x),
\qquad
y=E_{j^\star}(x).
\]
The read is single-shard but multi-skill: Address selects one capability
package, and the chosen executor can compose several canonical skills from that
package while resolving request-specific arguments and graph dependencies.
The language-model cost is one executor generation per request. Because Address
uses hashing plus Naive Bayes, shard selection requires neither an additional
language-model forward pass nor bank-wide scoring.

%% file: sections/06_main_results.tex
\section{Main Results}

\subsection{Experimental setup}

We evaluate whether continually internalized skills remain usable after later
skill families are acquired. Our primary experiments use Qwen3-8B with
rank-16 LoRA on the $q/k/v/o$ projections; Qwen3-4B provides an additional
model-scale robustness check. Each stage executor is trained for six epochs
with learning rate $4\times10^{-4}$ and effective batch size eight, then
frozen permanently. Evaluation uses greedy decoding with thinking disabled and
a 768-token generation cap. We report both task orders from
Section~\ref{sec:benchmark}. At test time, FSM receives only the raw request:
its replay-free Address Memory selects one stage-level execution shard, and
only that frozen executor generates the workflow.

\paragraph{Dataset.}
We use the TaskBench stream from Section~\ref{sec:benchmark}: 100 skills in
five disjoint families, 1,000 training requests per stage, and 2,186 held-out
requests. We evaluate $[T1,T2,T3,T4,T5]$ and
$[T3,T1,T5,T2,T4]$. Closed-book methods receive only the raw request; correct-skill-address
rows supply the correct skill address only for diagnostic evaluation.

\paragraph{Baselines.}
We include three reference conditions with additional information or memory.
\textbf{Base + Skill} gives the frozen base model the full skill catalog at
inference, \textbf{Base + Oracle Address+Fields} supplies the correct skill address and
gold argument fields, and \textbf{Replay} retains 100 examples from each previous
task. Our replay-free continual baselines are \textbf{LwF}
\citep{li2017learning}, \textbf{EWC} \citep{kirkpatrick2017overcoming},
\textbf{O-LoRA} \citep{wang2023orthogonal}, and
\textbf{Progressive-LoRA}. For LwF, EWC, and O-LoRA, the correct-skill-address rows reevaluate the same final checkpoints with the
correct skill address and no retraining; these rows are diagnostic rather than
deployment conditions.

\begin{table*}[!t]
\centering
\caption{Main comparison across both task orders. Metrics are on a 0--100
scale; Order~2 values are placed by task identity. Tool Choice is ordered
tool-list exact match, Arg.\ F1 compares argument-value tuples, Avg. is
task-macro EM, and BWT is final minus post-acquisition EM over the first four
arrival stages. Indented \texttt{w/ correct skill address} rows report
percentage-point changes from the closed-book row above. A dash denotes an
unreported metric.}
\label{tab:main-comparison}
\footnotesize
\setlength{\tabcolsep}{2.0pt}
\renewcommand{\arraystretch}{1.02}
\setlength{\heavyrulewidth}{0.11em}
\begin{tabular*}{\textwidth}{@{\extracolsep{\fill}}lrr@{\hspace{5pt}}rrrrrr@{\hspace{5pt}}r@{}}
\toprule
\multirow{2}{*}{Method} &
\multirow{2}{*}{Tool Choice} &
\multirow{2}{*}{Arg.\ F1} &
\multicolumn{6}{c}{EM} &
\multirow{2}{*}{BWT}\\
\cmidrule(lr){4-9}
& & & T1 & T2 & T3 & T4 & T5 & Avg. & \\
\midrule
Base + Skill & 49.57 & 52.24 & 19.91 & 22.79 & 50.30 & 29.77 & 47.70 & 34.09 & --\\
Base + Oracle Address+Fields & 90.71 & 81.01 & 23.84 & 70.10 & 59.94 & 69.23 & 56.89 & 56.00 & --\\
\midrule
\rowcolor{ordergray}
\multicolumn{3}{l}{\textbf{Order 1: $T1\rightarrow T2\rightarrow T3\rightarrow T4\rightarrow T5$}} &
\multicolumn{6}{c}{} & \\
Replay (reference) & 85.90 & 83.75 & 43.75 & 77.21 & 70.48 & 80.27 & 74.91 & 69.32 & -1.76\\
\midrule
Progressive-LoRA & 65.54 & 65.89 & 31.13 & 66.67 & 32.83 & 76.25 & 53.00 & 51.98 & -8.38\\
LwF~\citep{li2017learning} & 18.87 & 17.52 & 0.00 & 0.00 & 0.00 & 0.00 & \textbf{73.85} & 14.77 & -70.76\\
\hspace{0.9em}w/ correct skill address & +45.91 & +29.84 & +1.16 & +58.33 & +0.00 & +73.58 & +3.54 & +27.32 & --\\
EWC~\citep{kirkpatrick2017overcoming} & 9.89 & 10.53 & 0.00 & 0.00 & 0.00 & 0.00 & 44.88 & 8.98 & -54.94\\
\hspace{0.9em}w/ correct skill address & +21.73 & +19.35 & +0.00 & +0.00 & +0.00 & +67.56 & +6.71 & +14.85 & --\\
O-LoRA~\citep{wang2023orthogonal} & 18.45 & 17.53 & 0.00 & 0.00 & 0.00 & 0.00 & \textbf{73.85} & 14.77 & -70.19\\
\hspace{0.9em}w/ correct skill address & +70.79 & +50.42 & +1.16 & +81.13 & +70.18 & +72.24 & -2.47 & +44.45 & --\\
\midrule
\textbf{FSM} & \textbf{82.57} & \textbf{80.71} & \textbf{44.21} & \textbf{78.19} & \textbf{59.64} & \textbf{78.93} & 70.67 & \textbf{66.33} & \textbf{-3.71}\\
\specialrule{0.085em}{1.5pt}{1.5pt}
\rowcolor{ordergray}
\multicolumn{3}{l}{\textbf{Order 2: $T3\rightarrow T1\rightarrow T5\rightarrow T2\rightarrow T4$}} &
\multicolumn{6}{c}{} & \\
Replay (reference) & 81.70 & 81.03 & 42.94 & 74.02 & 68.67 & 78.26 & 68.20 & 66.42 & 17.98\\
\midrule
Progressive-LoRA & 64.72 & 65.21 & 31.13 & 66.67 & 31.93 & 75.59 & 51.24 & 51.31 & -11.51\\
LwF~\citep{li2017learning} & 17.73 & 17.50 & 0.00 & 0.00 & 0.00 & 78.60 & 0.00 & 15.72 & -19.05\\
\hspace{0.9em}w/ correct skill address & +63.67 & +46.54 & +0.00 & +74.51 & +59.34 & +3.67 & +59.36 & +39.38 & --\\
EWC~\citep{kirkpatrick2017overcoming} & 16.92 & 16.62 & 0.00 & 0.00 & 0.00 & 75.25 & 0.00 & 15.05 & -62.50\\
\hspace{0.9em}w/ correct skill address & +24.19 & +11.68 & +3.12 & +41.67 & +0.00 & +3.68 & +0.00 & +9.69 & --\\
O-LoRA~\citep{wang2023orthogonal} & 17.39 & 17.36 & 0.00 & 0.00 & 0.00 & \textbf{78.93} & 0.00 & 15.79 & -68.77\\
\hspace{0.9em}w/ correct skill address & +71.44 & +52.37 & +1.85 & +83.33 & +72.29 & +0.67 & +66.78 & +44.98 & --\\
\midrule
\textbf{FSM} & \textbf{81.78} & \textbf{79.94} & \textbf{44.68} & \textbf{78.19} & \textbf{57.83} & 77.59 & \textbf{69.61} & \textbf{65.58} & \textbf{-5.11}\\
\bottomrule
\end{tabular*}
\end{table*}

\subsection{Overall and per-task performance}

Table~\ref{tab:main-comparison} evaluates the final checkpoints under both
task orders. FSM averages 65.95\% task-macro EM, 14.3 points above
Progressive-LoRA and within 1.9 points of Replay. Its performance is also
stable to task order, reaching 66.33\% on Order~1 and 65.58\% on Order~2.
The correct-skill-address diagnostic recovers several otherwise weak
closed-book checkpoints, showing that low autonomous performance can coexist
with substantial execution under the correct skill address, as observed in
Section~\ref{sec:hierarchical-forgetting}.

LwF, EWC, and O-LoRA lose closed-book performance on most earlier tasks,
whereas FSM retains much more of the stream. Its BWT is $-3.71$ on Order~1
and $-5.11$ on Order~2, versus $-8.38$ and $-11.51$ for Progressive-LoRA.
On Order~2, the correct-skill-address diagnostic raises LwF and EWC to
55.10\% and 24.74\%, showing that some failures still preserve execution under
the correct skill address.

Table~\ref{tab:all-4b-order1} provides a Qwen3-4B robustness check. FSM
reaches 66.63\% task-macro EM with BWT $-3.58$, close to the Qwen3-8B
Order~1 result of 66.33\% and $-3.71$. Because archived 4B baselines use
different training recipes, this comparison is descriptive, but the similar
FSM result suggests that the factorized design is not specific to the 8B
backbone.

\begin{table*}[!t]
\centering
\caption{Qwen3-4B Order-1 results. Tool Choice and Arg.\ F1 are
percentages; T1--T5 report final closed-book graph exact match, Avg.\ is the
task-macro mean, and BWT is mean final-minus-acquisition EM over T1--T4.
Correct-skill-address rows are percentage-point deltas from the row above.}
\label{tab:all-4b-order1}
\footnotesize
\setlength{\tabcolsep}{2.0pt}
\renewcommand{\arraystretch}{1.02}
\setlength{\heavyrulewidth}{0.11em}
\begin{tabular*}{\textwidth}{@{\extracolsep{\fill}}lrr@{\hspace{5pt}}rrrrrr@{\hspace{5pt}}r@{}}
\toprule
\multirow{2}{*}{Method} &
\multirow{2}{*}{Tool Choice} &
\multirow{2}{*}{Arg.\ F1} &
\multicolumn{6}{c}{EM} &
\multirow{2}{*}{BWT}\\
\cmidrule(lr){4-9}
& & & T1 & T2 & T3 & T4 & T5 & Avg. & \\
\midrule
Replay (reference) & 86.58 & 84.05 & 43.87 & 80.15 & 71.08 & 79.26 & 74.20 & 69.71 & -1.39\\
\midrule
Progressive-LoRA & 56.78 & 60.03 & 18.29 & 62.99 & 29.82 & 73.58 & 49.82 & 46.90 & -7.84\\
LwF~\citep{li2017learning} & 18.80 & 17.53 & 0.00 & 0.00 & 0.00 & 0.00 & 73.85 & 14.77 & -70.31\\
\hspace{0.9em}w/ correct skill address & +17.73 & +15.21 & +0.00 & +0.00 & +0.00 & +70.23 & -0.35 & +13.98 & --\\
EWC~\citep{kirkpatrick2017overcoming} & 8.20 & 9.03 & 0.00 & 0.00 & 0.00 & 0.00 & 38.52 & 7.70 & -37.64\\
\hspace{0.9em}w/ correct skill address & -8.20 & -9.03 & +0.00 & +0.00 & +0.00 & +0.00 & -38.52 & -7.70 & --\\
O-LoRA~\citep{wang2023orthogonal} & 18.45 & 17.53 & 0.00 & 0.00 & 0.00 & 0.00 & 72.79 & 14.56 & -69.96\\
\hspace{0.9em}w/ correct skill address & +66.79 & +47.28 & +0.00 & +71.57 & +71.39 & +67.56 & +2.12 & +42.53 & --\\
\midrule
\textbf{FSM} & \textbf{82.60} & \textbf{80.48} & \textbf{44.79} & \textbf{79.17} & \textbf{57.53} & \textbf{79.93} & \textbf{71.73} & \textbf{66.63} & \textbf{-3.58}\\
\bottomrule
\end{tabular*}
\end{table*}

\paragraph{Selector variants.}
To characterize request-to-shard selection with a fixed execution bank, we
compare Oracle routing, confidence-based Self-route, the replay-free FSM
Address, EPI task identification \citep{wang2023epi}, and AdapterCL
perplexity routing \citep{madotto2021continual}.

\begin{table}[!t]
\centering
\caption{Per-task Graph EM for selector variants across both task orders.
EPI and AdapterCL use the same frozen execution banks as FSM.
Macro is the task average.}
\label{tab:per-task}
\small
\setlength{\tabcolsep}{4.0pt}
\renewcommand{\arraystretch}{0.96}
\begin{tabular}{llrrrrr}
\toprule
Order & Task & Oracle & Self-route & EPI & AdapterCL & FSM\\
\midrule
\multirow{6}{*}{\textbf{Order 1}}
 & T1 & 45.02 & 42.01 & 35.18 & 41.25 & \textbf{44.21}\\
 & T2 & 83.58 & 58.82 & 59.41 & 64.19 & \textbf{78.19}\\
 & T3 & 75.60 & 56.02 & 51.09 & \textbf{62.89} & 59.64\\
 & T4 & 80.27 & 77.26 & 63.94 & 70.74 & \textbf{78.93}\\
 & T5 & 76.33 & \textbf{72.44} & 54.84 & 64.24 & 70.67\\
 & Macro & 72.16 & 61.31 & 52.90 & 60.66 & \textbf{66.33}\\
\midrule
\multirow{6}{*}{\textbf{Order 2}}
 & T1 & 45.49 & 42.82 & 35.55 & 41.25 & \textbf{44.68}\\
 & T2 & 83.33 & 56.62 & 59.41 & 63.97 & \textbf{78.19}\\
 & T3 & 73.80 & 57.23 & 49.88 & \textbf{63.16} & 57.83\\
 & T4 & 78.93 & \textbf{77.93} & 62.88 & 70.74 & 77.59\\
 & T5 & 75.27 & \textbf{73.14} & 53.99 & 64.87 & 69.61\\
 & Macro & 71.36 & 61.55 & 52.34 & 60.80 & \textbf{65.58}\\
\bottomrule
\end{tabular}
\end{table}

\begin{figure}[!t]
  \centering
  \includegraphics[width=0.96\linewidth]{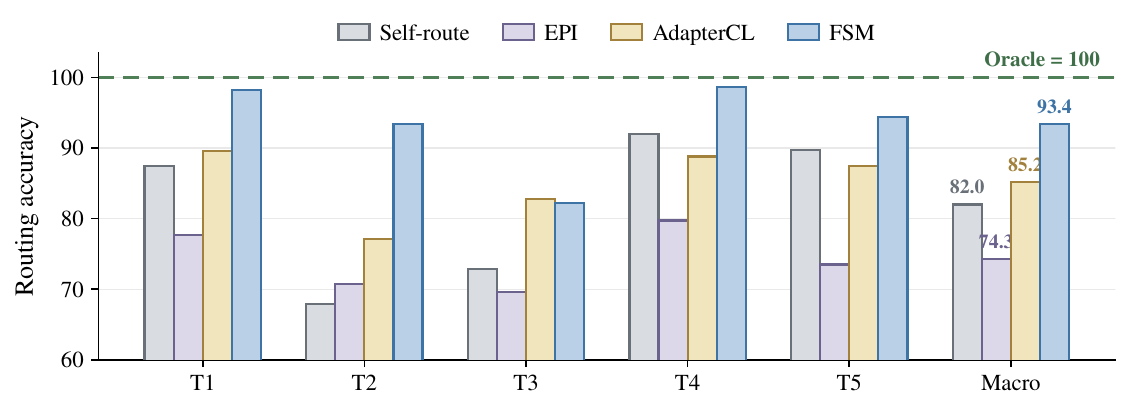}
  \caption{Two-order mean routing accuracy by task for Self-route, EPI,
  AdapterCL, and FSM. The dashed horizontal line marks Oracle routing at
  100\%. Macro is the task average.}
  \label{fig:fsm-selector-em}
\end{figure}

FSM reaches 93.38\% mean routing accuracy, compared with 85.15\% for
AdapterCL and 74.28\% for EPI; with the same frozen execution banks, their
mean Graph EM is 65.95\%, 60.73\%, and 52.62\%, respectively.
Figure~\ref{fig:fsm-selector-em} shows the task-level routing differences.

Oracle routing reaches 71.76\% mean EM, compared with 65.95\% for FSM and
61.43\% for Self-route. With the execution bank fixed, the 5.81-point
FSM--Oracle gap reflects remaining shard-access error. Oracle itself still
lies 28.24 points below perfect EM after the correct shard is supplied,
because the selected executor must recover the skill sequence, arguments,
dependencies, and complete workflow. Thus both access and within-shard
execution contribute materially to the final error.

The task-level pattern makes this split concrete. T2 is mostly access-limited:
Oracle reaches 83.58\% / 83.33\%, FSM 78.19\% in both orders, and
Self-route 58.82\% / 56.62\%. T1 is mostly execution-limited because FSM
nearly matches an Oracle that reaches only 45.02\% / 45.49\%. T3 is mixed:
Oracle remains above 73\%, while all learned selectors leave a larger gap;
AdapterCL is stronger than FSM on this family. T4 again lies close to Oracle,
whereas Self-route slightly exceeds FSM on T5. These differences show why a
single final EM score is insufficient: some families are dominated by access
error and others by the execution quality of the selected shard. FSM's macro
advantage comes mainly from recovering the access-sensitive cases while
remaining competitive elsewhere.

With the execution bank fixed, offline Address reaches 69.15\% mean EM and
97.03\% routing accuracy, versus 65.95\% and 93.38\% for sequential Address
(Appendix~\ref{app:representation-calibration}). This 3.20-point difference
serves as a historical-request reference rather than an isolated replay effect,
because the two selectors also differ in both representation and classifier
family (hashing+NB versus TF--IDF+LR). The stronger offline selector closes a noticeable portion of the remaining
Oracle gap, indicating that improved request-to-shard access still offers
meaningful headroom beyond the deployed sequential Address.

FSM avoids bank-wide language-model scoring: Address performs a lightweight
request-to-shard lookup, activates one frozen LoRA, and uses only one executor
generation for each incoming request. Each rank-16 execution shard adds 15.34M
parameters; five shards add 76.68M, or only 0.94\% of the shared 8.2B
backbone. The backbone itself is stored only once, and only the selected LoRA
is activated during inference. As Execution Memory grows, storage therefore
increases with the number of frozen shards, while the per-request
language-model cost remains a single executor generation.

%% file: sections/10_conclusion.tex
\section{Conclusion}
Correct-skill-address reactivation reveals a hierarchical forgetting pattern:
closed-book execution can collapse while execution under the correct skill
address remains substantially recoverable, and continued optimization can
later erase that recoverability as well. FSM follows this distinction by
keeping Address Memory plastic and storing acquired procedures in frozen,
multi-skill Execution Memory shards that remain unchanged during later stages.
Across two five-stage task orders, FSM reaches 65.95
93.38
percentage points while remaining within 1.9 points of Replay. These results
support a simple design principle for continual skill memory: preserve acquired
procedural content while allowing access to adapt as the skill stream grows.
They also separate two remaining sources of error: Address determines which
execution shard is reached, whereas the selected shard determines how well the
requested workflow is executed. This decomposition makes retention failures
more interpretable and suggests where additional capacity or supervision is
most useful.

%% file: sections/appendix_a_reproducibility.tex
\section{Experimental details and reproducibility}
\label{app:benchmark-metrics}
This appendix records the benchmark protocol, evaluation contract, metrics,
FSM implementation, information budgets, and safeguards used for the main
results.

\subsection{Continual internalization protocol}
At stage $t$, the learner receives request--graph pairs from a previously unseen
skill set $\mathcal S_t$ and then moves to $\mathcal S_{t+1}$. The skill sets
are pairwise disjoint, and earlier skills remain part of the target competence.
For a request $x$, the model must produce an executable graph $g$: nodes specify
canonical skills and arguments, and directed edges encode producer--consumer
dependencies. Training may use graph supervision and, for skill-guided
baselines, textual skill cards. Evaluation is closed-book: the model receives
only $x$, with no stage label, skill name, catalog, demonstration, retrieved
memory, or target graph.

We evaluate two complete task orders,
$O1=[T1,T2,T3,T4,T5]$ and $O2=[T3,T1,T5,T2,T4]$. The primary metric is exact
match of the canonical skill graph. Component metrics separate errors in skill
selection, argument values, and dependency edges. Backward transfer measures
how performance on earlier tasks changes after later stages. All reported
aggregates are task macros, so each family has equal weight despite different
held-out set sizes.

\subsection{Benchmark inventory and closed-book contract}
Table~\ref{tab:benchmark-inventory} gives the five-family split. Each family
contains 1,000 training requests, and final task scores use its complete
held-out set. The headline aggregate is the macro mean across families.

\begin{table}[H]
\centering
\scriptsize
\begin{tabular}{llrrr}
\toprule
ID & Family & Skills & Train & Held-out\\
\midrule
T1 & Daily Life & 40 & 1,000 & 864\\
T2 & Language/Speech & 11 & 1,000 & 408\\
T3 & Retrieval/Understanding & 21 & 1,000 & 332\\
T4 & Vision/Document & 9 & 1,000 & 299\\
T5 & Generation/Editing & 19 & 1,000 & 283\\
\midrule
Total & Five disjoint families & 100 & 5,000 & 2,186\\
\bottomrule
\end{tabular}
\caption{Benchmark inventory. Train and held-out counts are request--graph
pairs; the two orders reuse the same held-out sets.}
\label{tab:benchmark-inventory}
\end{table}

The closed-book system contract requires exactly one JSON object containing
\texttt{task\_nodes} and \texttt{task\_links}. Nodes must be topologically
ordered and include a canonical task identifier with named arguments. Links
specify directed producer--consumer dependencies. The user message contains
only the original request; skill names, family IDs, catalogs, workflows,
demonstrations, retrieval results, and target graphs are absent. We use greedy
generation with thinking disabled and a 768-token cap. Evaluation parses and
canonicalizes the response before computing graph and component metrics. No
test-informed repair is applied.

\subsection{Closed-book prompt and an example record}
Figure~\ref{fig:taskbench-prompt} reproduces the system instruction and user
template used for all five stages. The model uses its chat template with
thinking disabled. Gold graphs are training targets and evaluation references,
never closed-book inputs. Both task orders use the same prompt and held-out
requests; only the acquisition order changes.

\begin{figure}[!t]
\centering
\begin{minipage}{0.98\linewidth}
\begin{lstlisting}[basicstyle=\ttfamily\footnotesize,breaklines=true,columns=fullflexible,keepspaces=true,frame=single]
[system]
You are a task-automation planner.
Return exactly one JSON object and no other text.
The JSON object must have exactly these two keys in this order: task_nodes, task_links.
task_nodes is a non-empty topologically ordered list. Every node has exactly task and arguments.
task is an exact canonical skill identifier. arguments is a list of objects with exactly name and value.
task_links is a list of objects with exactly source and target. Use <node-i> when an argument consumes node i.
Use the minimum skills needed. Do not emit markdown, explanation, a skill list, a plan, or text before or after JSON.

[user]
Plan the following request:
{user_request}
\end{lstlisting}
\end{minipage}
\caption{Closed-book TaskBench prompt template. The system instruction is
reproduced from the benchmark records; the user request is substituted into
the final line. No skill card, gold skill name, family label, or target graph
appears in the input.}
\label{fig:taskbench-prompt}
\end{figure}
\FloatBarrier

\noindent\begin{minipage}{\linewidth}
\paragraph{Illustrative training pair.}
The following T1 Daily Life training record requires two skills and one dependency link.
The graph is pretty-printed for readability; the original target is the same
JSON object in compact form.
\begin{lstlisting}[basicstyle=\ttfamily\footnotesize,breaklines=true,columns=fullflexible,keepspaces=true,frame=single,aboveskip=0.35em,belowskip=0pt]
Request: I want to pay my electricity bill and then, attend an online meeting about 'Electricity Bill Management'.

Gold graph:
{"task_nodes": [
  {"task": "TaskBench::t1_daily_life::daily_bill_payment",
   "arguments": [{"name": "bill", "value": "electricity"}]},
  {"task": "TaskBench::t1_daily_life::attend_meeting_online",
   "arguments": [{"name": "topic", "value": "Electricity Bill Management"}]}
 ],
 "task_links": [
  {"source": "TaskBench::t1_daily_life::daily_bill_payment",
   "target": "TaskBench::t1_daily_life::attend_meeting_online"}
 ]}
\end{lstlisting}
\end{minipage}

\subsection{Metric definitions}
For a parsed prediction $\hat g=(\hat V,\hat E,\hat\phi)$ and target
$g=(V,E,\phi)$, Graph EM is
\[
  \mathbf{1}[\operatorname{canon}(\hat g)=\operatorname{canon}(g)].
\]
Tool-choice accuracy is exact match of the ordered tool-name sequence.
Argument-value F1 compares multisets of task, argument-name, and argument-value
tuples. These diagnostics distinguish choosing the correct procedure from
filling its arguments correctly.

For finer analysis, node-set exactness compares the canonical node multiset,
argument-value exactness requires exact equality of the value-tuple multiset,
and edge exactness compares the directed edge multiset. A prediction can match
the nodes and still fail Graph EM because of one incorrect value or link. Edge
exactness is not a standalone success measure: single-node targets have no gold
edges, so an empty predicted edge set can be edge-exact even when the node is
wrong. Route accuracy is
\[
  |D|^{-1}\sum_{(x,j)\in D}\mathbf{1}[\hat j(x)=j],
\]
which measures executor selection rather than graph correctness. It is distinct
from tool-choice accuracy within the generated graph. For continual accuracy
matrix $a_{t,k}$, backward transfer is
\[
  \operatorname{BWT}=\frac{1}{T-1}\sum_{k=1}^{T-1}
  (a_{T,k}-a_{k,k}),
\]
where negative values indicate forgetting relative to performance immediately
after task $k$ was learned. Task-macro averages assign equal weight to all five
families.

\subsection{FSM implementation and information budgets}
\begin{table}[!t]
\centering
\scriptsize
\begin{tabular}{p{0.29\columnwidth}p{0.57\columnwidth}}
\toprule
Reproducibility item & Frozen definition in the current candidate\\
\midrule
Executor checkpoint & Rank-16 LoRA on $q/k/v/o$, learning rate $4\times10^{-4}$, six epochs, epoch 6 exported.\\
Execution-memory storage & Shared 8.2B backbone plus 15.34M LoRA parameters per shard; five shards add 76.68M parameters ($0.94\%$ of the backbone).\\
Sequential Address & Request text only; fixed-dimensional hashing features plus incremental Naive Bayes; update on current stage only.\\
Retained Address state & Compact class/feature statistics; no old requests or workflow targets are replayed.\\
Inference & Hard request-level routing to one frozen executor; one language-model generation per request.\\
Reference selector & Offline TF--IDF + logistic regression uses all historical requests.\\
\bottomrule
\end{tabular}
\caption{Protocol choices that define the deployed FSM method and its
reference variants.}
\label{tab:reproducibility}
\end{table}

\begin{algorithm}[!t]
\caption{Factorized Skill Memory: replay-free memory update and read}
\label{alg:fsm}
\begin{algorithmic}[1]
\Require Frozen base model $B$; stage data $\{\mathcal D_t\}_{t=1}^{T}$
\State $\mathcal E_0\gets\varnothing$; initialize Address Memory $A_0$
\For{$t=1,\ldots,T$}
    \State Initialize a fresh LoRA $\Delta_t$ on the frozen base $B$
    \State Train $\Delta_t$ using only current-stage supervision $\mathcal D_t$
    \State $E_t\gets B\oplus\Delta_t$; freeze $E_t$
    \State $\mathcal E_t\gets\mathcal E_{t-1}\cup\{E_t\}$
    \State $A_t\gets\operatorname{UpdateAddress}(A_{t-1},
           \mathrm{Req}(\mathcal D_t),t)$
    \State Discard the stage-$t$ requests after updating $A_t$
\EndFor
\Statex
\Function{Infer}{$x$}
    \State $p_j\gets p_{\mathrm{Address}}(j\mid x;A_T)$ for $j=1,\ldots,T$
    \State $j^\star\gets\arg\max_j p_j$
    \State \Return $\operatorname{Generate}(E_{j^\star},x)$
\EndFunction
\end{algorithmic}
\end{algorithm}

Each executor is fixed after its stage finishes. Address is then updated without
replaying old requests. Qwen3-8B has 36 transformer layers with hidden size
4096, 32 query heads, and 8 key/value heads \citep{yang2025qwen3}. For rank
$r=16$, the LoRA tensors
on $q/k/v/o$ contribute
$36r[(4096+4096)+2(4096+1024)+(4096+4096)]=15{,}335{,}424$
parameters per shard. Five shards therefore contribute 76.68M parameters, or
about 146 MiB when stored as BF16 tensors. Hard routing activates one shard per
request, so deployed inference still uses one language-model generation.

\subsection{Information budgets and comparator roles}
The selectors differ in the information they receive. Oracle uses the source
family label to measure conditional execution. Self-route sees executor
generations and token scores but no family label. FSM receives the
current request pool during training, retains only classifier statistics, and
runs one selected executor at test time. FSM-Offline revisits the full historical request set. Replay uses old
request--workflow pairs during training.

These budgets define the comparisons: Oracle--FSM includes residual addressing
and executor error. FSM-Offline--FSM is a historical-request reference rather
than an isolated replay ablation, because the two Address variants also use
different feature representations and classifiers.

\subsection{Evaluation safeguards and artifact provenance}
The evaluation packet contains 50 rollout files across the two task orders,
with 25 task--executor cells per order. Address uses only request text at
inference; family labels are training supervision only, and held-out graphs are
not used for model or hyperparameter selection. Each rollout is checked for row
count, closed-book status, the 768-token cap, and token-score conventions.
Separate result summaries track learned Address, Sequential Address, semantic
controls, and template-group controls. Self-route additionally retains
per-request candidate indices for the coverage--selection analysis in
Appendix~\ref{app:audits}.

%% file: sections/appendix_d_additional_audits.tex
\section{Additional analysis and controls}
\label{app:audits}
These controls test the access--execution split, alternative Address
representations, and routing as the executor bank grows.

\subsection{Correct-skill-address reactivation controls}
\label{app:skill-name-controls}
At the frozen T2 checkpoint used in the main text, closed-book full-task EM is
0\% over 408 requests and rises to 69.85\% when the correct skill address is
supplied. The address contains no request-specific values, intermediate
results, workflow steps, demonstration, or target serialization; the model
still reconstructs the request-conditioned workflow. Across all requests,
evaluation under the correct skill address reaches 70.10\% argument-value
exactness and 69.85\% full-task EM.

The recovery is specific to the correct skill address. An incorrect skill
address from the same family yields 6.86\% full-task EM, while a semantic
description, an unrelated valid address, token-matched random text, and a
generic instruction all remain at 0\%. These controls rule out generic
non-target cues, but the correct skill address still reveals target node
identity.

\paragraph{Frozen-base correct-skill-address control.}
Table~\ref{tab:base-name-only-control} evaluates frozen Qwen3-8B on all 2,186
held-out requests with only the correct skill address. No argument-field
names, workflow, or answer are supplied.

\begin{table}[!t]
\centering
\caption{Frozen Qwen3-8B with the correct skill address only (percent).}
\label{tab:base-name-only-control}
\scriptsize
\begin{tabular}{lrrr}
\toprule
Task & Requests & Full-task EM & Skill selection\\
\midrule
T1 & 864 & 7.87 & 83.22\\
T2 & 408 & 0.00 & 92.65\\
T3 & 332 & 0.00 & 87.05\\
T4 & 299 & 0.00 & 93.65\\
T5 & 283 & 0.00 & 94.35\\
\midrule
Macro & -- & \textbf{1.57} & --\\
\bottomrule
\end{tabular}
\end{table}

The base model identifies the intended skill on 87.05--94.35\% of T2--T5
requests but obtains 0\% full-task EM on all four families; T1 reaches 7.87\%
EM. Address recognition alone therefore does not explain the T2 reactivation
result.

\subsection{Coverage--selection decomposition}
Table~\ref{tab:decomp} separates self-routing coverage from selection. $O=C$
marks requests for which the matching executor is correct and at least one
candidate in the bank is correct; $C-R$ is the loss from confidence-based
selection among those candidates.

\begin{table}[!t]
\centering
\scriptsize
\caption{Current self-routing coverage and selection decomposition (percent).
The table applies only to self-routing; learned Address did not retain
per-request candidate indices.}
\label{tab:decomp}
\begin{tabular}{lrrrr}
\toprule
Order/task & $O=C$ & Self-route $R$ & All wrong & $C-R$\\
\midrule
O1/T1 & 45.02 & 42.01 & 54.98 & 3.01\\
O1/T2 & 83.58 & 58.82 & 16.42 & 24.75\\
O1/T3 & 75.60 & 56.02 & 24.40 & 19.58\\
O1/T4 & 80.27 & 77.26 & 19.73 & 3.01\\
O1/T5 & 76.33 & 72.44 & 23.67 & 3.89\\
O2/T1 & 45.49 & 42.82 & 54.51 & 2.66\\
O2/T2 & 83.33 & 56.62 & 16.67 & 26.72\\
O2/T3 & 73.80 & 57.23 & 26.20 & 16.57\\
O2/T4 & 78.93 & 77.93 & 21.07 & 1.00\\
O2/T5 & 75.27 & 73.14 & 24.73 & 2.12\\
\bottomrule
\end{tabular}
\end{table}

\subsection{Evidence for the access--execution split}
Table~\ref{tab:claim-audit} aligns each intervention with the part of the
pipeline it measures: name cues and Oracle expose conditional execution,
whereas Self-route and Address test recovery of the appropriate executor from
the raw request.

\begin{table}[!t]
\centering
\scriptsize
\begin{tabular}{p{0.24\columnwidth}p{0.39\columnwidth}p{0.25\columnwidth}}
\toprule
Question & Observed evidence & What it measures\\
\midrule
Is execution retained? & Oracle Graph EM is 71.76\% across two orders. & Conditional execution under perfect shard access.\\
Is access unreliable? & Self-route is 61.43\% while candidate coverage is 71.76\%. & Selection loss is measured only for self-route.\\
Can request text recover access continually? & FSM reaches 65.95\% and 93.38\% route accuracy. & Address itself remains sequence-sensitive.\\
What is available with historical requests? & Offline Address reaches 69.15\%. & Reference selector with historical-request access.\\
\bottomrule
\end{tabular}
\caption{Measurements supporting the Address--Execution decomposition.}
\label{tab:claim-audit}
\end{table}

\subsection{Factorized error decomposition}
Oracle, Self-route, and FSM separate executor error from access error. Oracle
uses the matching executor, Self-route uses generation confidence, and FSM
learns request-to-shard access under the replay-free sequential constraint.

\begin{table}[!t]
\centering
\scriptsize
\begin{tabular}{p{0.24\columnwidth}p{0.31\columnwidth}p{0.31\columnwidth}}
\toprule
Observed pattern & Primary diagnosis & Next measurement\\
\midrule
Low oracle and low routing & Conditional execution is weak. & Revisit stage training, capacity, or data before tuning Address.\\
High oracle and low self-route & Usable skills exist but are selected unreliably. & Compare request-only routing rules.\\
Sequential below offline & Historical-request reference is stronger, but the cause is confounded by selector design. & Match representation and classifier under online/offline updates.\\
Offline close to oracle & A stronger historical-request selector leaves a smaller access gap. & Separate update regime from representation/classifier choice.\\
\bottomrule
\end{tabular}
\caption{Error decomposition induced by Oracle, Self-route, and learned Address.}
\label{tab:diagnostic-guide}
\end{table}

\subsection{Representation and calibration controls}
\label{app:representation-calibration}

Table~\ref{tab:controls} compares replay-free Address with offline and
representation controls. Because the deployed sequential Address uses
hashing plus Naive Bayes while the offline variants use different
representations and classifiers, these rows are reference comparisons rather
than isolated replay ablations. The template-group split removes exact value,
filename, and number matches before fitting the offline classifier.

\begin{table}[!t]
\centering
\caption{Representation and routing controls (percent). Training-request
classification and final-test route accuracy are distinct quantities.}
\label{tab:controls}
\scriptsize
\begin{tabular}{lrrl}
\toprule
Address/control & Route acc. & EM & Evaluation scope\\
\midrule
FSM & 93.38 & 65.95 & replay-free, two-order mean\\
Lexical TF--IDF (offline) & 97.03 & 69.15 & historical requests retained\\
E5-base-v2 & 94.22 & 66.61 & final test, two-order mean\\
Qwen3-4B mean-pool & 90.92 & 64.20 & final test, two-order mean\\
Template-group split & 97.09 & -- & 997 held-out train rows\\
Ordinary five-fold split & 97.26 & -- & training requests\\
\bottomrule
\end{tabular}
\end{table}

\subsection{Training sensitivity check}
\label{app:training-sensitivity}

Table~\ref{tab:tuning-screen} reports a shared T1/T5 screen over learning
rate, batch size, and epoch. Both model sizes select $4\times10^{-4}$ with
batch eight; the 8B batch-32 condition is worse.

\begin{table}[!t]
\centering
\caption{Controlled T1/T5 validation screen (percent). The screen used
800/200 training/validation rows per task and epochs 1--3.}
\label{tab:tuning-screen}
\scriptsize
\begin{tabular}{lrrrr}
\toprule
Task & 4B best & 8B best & 8B batch-32 & Selected recipe\\
\midrule
T1 EM & 47.5 & 44.5 & 32.5 & $4\times10^{-4}$, batch 8\\
T5 EM & 77.5 & 76.5 & 71.0 & $4\times10^{-4}$, batch 8\\
\bottomrule
\end{tabular}
\end{table}

The controls separate routing from execution: frozen representations remain
below the lexical Address, and route accuracy exceeds end-to-end EM when the
selected executor still makes execution errors.

\paragraph{Model-scale and representation robustness.}
The 8B oracle average is 71.76\%, close to the historical 4B value of
71.94\%; the cross-scale comparison is descriptive because the training
recipes differ. In the controlled T1/T5 screen, both sizes select the same
learning rate and batch size, with 4B/8B validation EM of 47.5/44.5 on T1 and
77.5/76.5 on T5. E5-base-v2 and Qwen3-4B mean-pool representations remain
below the sparse lexical Address, and the template-group split reaches 97.09\%
Address accuracy versus 97.26\% under ordinary five-fold splitting.

\subsection{Sequential Address prefixes}
Table~\ref{tab:seq-prefix} tracks Sequential Address after every prefix.
Final route accuracy is 93.38\% in both orders; O1 drops most at T2--T3,
whereas O2 drops more after T3--T5. The executor bank is fixed throughout.

\begin{table}[!t]
\centering
\scriptsize
\begin{tabular}{lrrrr}
\toprule
Order prefix & Oracle & FSM & Route accuracy & Available tasks\\
\midrule
O1/$n=1$ & 45.02 & 45.02 & 100.00 & T1\\
O1/$n=2$ & 64.30 & 63.77 & 98.88 & T1--T2\\
O1/$n=3$ & 68.07 & 63.59 & 94.30 & T1--T3\\
O1/$n=4$ & 71.12 & 66.70 & 94.67 & T1--T4\\
O1/$n=5$ & 72.16 & 66.33 & 93.38 & T1--T5\\
\midrule
O2/$n=1$ & 73.80 & 73.80 & 100.00 & T3\\
O2/$n=2$ & 59.64 & 59.41 & 99.39 & T3--T1\\
O2/$n=3$ & 64.85 & 62.10 & 96.64 & T3--T1--T5\\
O2/$n=4$ & 69.47 & 63.19 & 92.73 & T3--T1--T5--T2\\
O2/$n=5$ & 71.36 & 65.58 & 93.38 & T3--T1--T5--T2--T4\\
\bottomrule
\end{tabular}
\caption{Sequential Address prefix results (percent). Oracle is the
conditional reference at each prefix; route accuracy is the replay-free
Address route audit.}
\label{tab:seq-prefix}
\end{table}

\subsection{Self-route prefix trajectories}
Table~\ref{tab:self-prefix} tracks confidence-based routing as the frozen bank
grows. Each row evaluates the tasks available at that prefix; Table~\ref{tab:seq-prefix}
gives the corresponding learned-Address trajectory.

\begin{table}[!t]
\centering
\caption{Current self-confidence Graph EM by order prefix (percent). Dashes
mark task families that had not entered the stream.}
\label{tab:self-prefix}
\scriptsize
\setlength{\tabcolsep}{2.7pt}
\renewcommand{\arraystretch}{0.96}
\begin{tabular}{@{}llrrrrr@{}}
\toprule
Order & Prefix & T1 & T2 & T3 & T4 & T5\\
\midrule
\multirow{5}{*}{O1}
 & $n=1$ & 45.02 & -- & -- & -- & --\\
 & $n=2$ & 42.82 & 82.60 & -- & -- & --\\
 & $n=3$ & 42.36 & 72.06 & 61.45 & -- & --\\
 & $n=4$ & 42.01 & 64.71 & 60.54 & 77.93 & --\\
 & $n=5$ & 42.01 & 58.82 & 56.02 & 77.26 & 72.44\\
\midrule
\multirow{5}{*}{O2}
 & $n=1$ & -- & -- & 73.80 & -- & --\\
 & $n=2$ & 44.33 & -- & 72.59 & -- & --\\
 & $n=3$ & 43.52 & -- & 68.07 & -- & 74.20\\
 & $n=4$ & 43.06 & 64.22 & 58.43 & -- & 73.50\\
 & $n=5$ & 42.82 & 56.62 & 57.23 & 77.93 & 73.14\\
\bottomrule
\end{tabular}
\end{table}

\subsection{Historical prefix and prompt controls}
The earlier Qwen3-4B one-order audit produced routed task rows of 44.68 at
stage 1; 44.21/83.09 at stage 2; 43.87/72.79/62.35 at stage 3;
43.87/68.14/60.84/78.60 at stage 4; and
43.87/61.52/57.23/77.93/71.73 at stage 5. In the same protocol, five
prompt-order variants reached final Graph EM of 60.28, 61.51, 59.55, 61.22,
and 59.67, while request-only self-routing reached 62.45. None of the prompt
variants exceeded request-only self-routing.

%% file: sections/08_analysis_limitations.tex
\section{Training dynamics and additional analysis}
\label{app:attempts}
This appendix records the training dynamics and capacity analyses behind the
factorized design.

\subsection{Memory growth and modularity}
Freezing executors prevents later stages from overwriting their parameters and
makes conditional retention directly measurable. The bank therefore grows with
the number of stages, as in other progressive and expert-bank methods
\citep{rusu2016progressive,aljundi2017expertgate}. Address is updated without
old requests or graphs and retains compact class statistics; deployed FSM
selects one executor before a single language-model generation.

\subsection{Training branches that motivate factorization}
Table~\ref{tab:attempt-inventory} summarizes the completed branches that led
to the final factorization.

\begin{table*}[!t]
\centering
\scriptsize
\begin{tabular}{p{0.22\textwidth}p{0.18\textwidth}p{0.50\textwidth}}
\toprule
Branch & Evidence level & Outcome and disposition\\
\midrule
Current-only SFT, GRPO/HC-GRPO, replay, LwF, EWC, OPSD, and adapter variants
& Complete five-task endpoints & Most direct updates overwrite old tasks; representative endpoints are reported in the main results and the analyses below.\\
SKILL0 TaskBench curriculum & Complete lower triangles & Current-task acquisition succeeds, but request-only old-task Graph EM collapses; exact names temporarily reactivate intermediate checkpoints.\\
One-epoch SFT warm start before GRPO/SKILL0 & Completed audited cells & Warm start enables current-task learning but does not prevent cross-task overwrite.\\
Pure on-policy SKILL0 pilot & Completed T1 pilot & Protocol assistance creates a weak learning signal, but strict closed-book Graph EM remains zero.\\
No-LoRA skill-information controls & Complete 30-cell grid & Canonical skill metadata exposes latent capability but changes the inference information boundary.\\
T2$\rightarrow$T3 learning-rate sweep & Complete 8B checkpoint grid & Lower learning rates delay the switch but do not yield a useful retention--plasticity operating point.\\
Frozen executors plus sequential Address & Complete two-order result & Promoted to FSM: execution is frozen and Address is updated without old-request replay.\\
\bottomrule
\end{tabular}
\caption{Training branches and the evidence they provide for the final design.}
\label{tab:attempt-inventory}
\end{table*}

\subsection{Sequential SFT, GRPO, and SKILL0 acquisition}

All three direct sequential recipes learn the current family and lose earlier
families. Table~\ref{tab:attempt-lower-triangle-summary} reports acquisition
on the diagonal and the final closed-book endpoint. SFT and Dense GRPO use
Qwen3-8B; the SKILL0 trajectory uses Qwen3-4B.

\begin{table*}[!t]
\centering
\scriptsize
\begin{tabular}{lrrrrrrl}
\toprule
Recipe & S1/T1 & S2/T2 & S3/T3 & S4/T4 & S5/T5 & Old tasks after S5 & Prompt\\
\midrule
Current-only SFT (8B) & 45.02 & 83.58 & 75.90 & 80.60 & 76.33 & all 0 & request only\\
GRPO (8B) & 40.86 & 79.90 & 73.19 & 79.93 & 71.73 & all 0 & request only\\
SKILL0 & 46.06 & 83.33 & 77.11 & 82.27 & 79.15 & all 0 & request only\\
SKILL0 & 46.88 & 83.09 & 76.51 & 80.60 & 79.86 & all 0 & full skill card\\
\bottomrule
\end{tabular}
\caption{Closed-book or supplied-skill Graph EM (percent) on the acquired
task at each stage. The final old-task statement is evaluated after S5. The
4B SKILL0 rows are not capacity-matched to the 8B baselines.}
\label{tab:attempt-lower-triangle-summary}
\end{table*}

Supplying the full skill card creates only a temporary retention window:
T2 reaches 58.33\% at S3 and 0\% at S4, with all old tasks at 0\% after S5.
The correct-skill-address audit follows the same pattern, reaching 52.94\% at S3 and
0\% at S4.

\subsection{Full forgetting trajectories and correct-skill-address reactivation}

The extended T2 audit evaluates every even optimizer step from 0 through 126
under request-only and correct-skill-address conditions. SFT and GRPO use Qwen3-8B
checkpoints; the SKILL0 trajectory uses Qwen3-4B. The curves compare the timing
of access and execution failure.

\begin{table}[!t]
\centering
\scriptsize
\begin{tabular}{lrrr}
\toprule
Method & First request-only zero & Skill address at step 126 & Request-only at 126\\
\midrule
SFT-8B & 8 & 54.66 & 0.00\\
GRPO-8B & 6 & 37.01 & 0.00\\
SKILL0-4B & 18 & 82.84 & 0.00\\
\bottomrule
\end{tabular}
\caption{Extended T2 retention trajectories (Graph EM, percent). Supplying the
correct skill address continues to recover behavior long after autonomous
closed-book access reaches zero.}
\label{tab:attempt-full-forgetting-curves}
\end{table}

Figure~\ref{fig:two-stage-forgetting} shows the early transition;
Table~\ref{tab:attempt-full-forgetting-curves} extends it to step 126.

\subsection{Warm starts and pure on-policy acquisition}

One SFT epoch produces a strong current-task policy. Warm-started GRPO moves
T1 from 39.93\% after SFT to 41.20\% after one GRPO epoch. In the warm-started
SKILL0 five-task sequence, final request-only scores are 0/0/0/0/72.44 versus
0/0/0/0/69.61 for one-epoch current-only SFT: the final task improves slightly
while old-task retention remains zero. In the zero-SFT on-policy T1 pilot,
direct skill-free training yields constant within-group rewards. A staged
protocol raises recall Graph EM from 4.17\% to 6.94\% after 40 steps and
7.18\% after 100 steps, with 22.34\% selection exactness, 52.39\% node F1,
and 28.21\% argument-value F1 at step 100. Strict closed-book Graph EM remains
0.

\subsection{Learning-rate sweep from the post-T2 checkpoint}
\label{app:lr-sweep}

We test whether a lower T3 learning rate can preserve T2 under current-only
SFT. Every arm starts from the same post-T2 Qwen3-8B checkpoint (T2 Graph EM
77.45\%), trains on 1,000 T3 examples for one epoch (125 optimizer steps,
global batch eight), uses no old-task replay, and follows the same held-out
request-only evaluation protocol.

\begin{table*}[!t]
\centering
\scriptsize
\resizebox{\textwidth}{!}{%
\begin{tabular}{lrrrrl}
\toprule
Learning rate & Best observed step & T2 Graph EM & T3 Graph EM & T2--T3 mean & Outcome\\
\midrule
$2\times10^{-5}$ & 100 & 0.00 & 74.40 & 37.20 & T3 learned; T2 erased\\
$1\times10^{-5}$ & 125 & 0.00 & 75.60 & 37.80 & T3 learned; T2 erased\\
$5\times10^{-6}$ & 50 & 0.00 & 76.81 & 38.40 & best T3; T2 erased\\
$2\times10^{-6}$ & 125 & 0.00 & 73.19 & 36.60 & T3 learned; T2 erased\\
$1\times10^{-6}$ & 25 & 79.17 & 0.00 & 39.58 & T2 retained before switch\\
$5\times10^{-7}$ & 50 & 78.43 & 0.00 & 39.22 & T2 retained before switch\\
$2\times10^{-7}$ & 75 & 78.19 & 0.00 & 39.09 & T3 never acquired\\
$1\times10^{-7}$ & 50 & 77.94 & 0.00 & 38.97 & T3 never acquired\\
\bottomrule
\end{tabular}
}
\caption{Best two-task mean observed in each learning-rate arm (percent).
For the four higher rates, every checkpoint with nonzero T3 has zero T2. For
the two lowest rates, T3 remains zero for the full epoch.}
\label{tab:lr-sweep-overview}
\end{table*}

The coarse grid shows an abrupt transition between retaining T2 and acquiring
T3. We evaluate checkpoints every five steps for the two learning rates nearest
that transition.

\begin{table}[!t]
\centering
\scriptsize
\begin{tabular}{lrrrrrr}
\toprule
LR & Step & T2 Graph EM & T3 Graph EM & Mean & T2 node F1 & T3 node F1\\
\midrule
$1\times10^{-6}$ & 25 & 79.17 & 0.00 & 39.58 & 94.45 & 0.00\\
& 30 & 79.41 & 0.00 & 39.71 & 95.45 & 0.00\\
& 35 & 0.00 & 18.98 & 9.49 & 0.00 & 28.22\\
& 40 & 0.00 & 39.76 & 19.88 & 0.00 & 56.35\\
& 45 & 0.00 & 51.20 & 25.60 & 0.00 & 70.15\\
& 50 & 0.00 & 57.23 & 28.61 & 0.00 & 77.34\\
\midrule
$5\times10^{-7}$ & 50 & 78.43 & 0.00 & 39.22 & 94.35 & 0.00\\
& 55 & 78.19 & 0.00 & 39.09 & 94.56 & 0.00\\
& 60 & 67.89 & 2.11 & 35.00 & 81.23 & 2.72\\
& 65 & 0.25 & 10.24 & 5.24 & 0.25 & 16.94\\
& 70 & 0.00 & 12.65 & 6.33 & 0.00 & 19.86\\
& 75 & 0.00 & 15.36 & 7.68 & 0.00 & 24.24\\
\bottomrule
\end{tabular}
\caption{Five-step zoom around the learning-rate transition (percent). The
two overlapping checkpoints occur before T3 is meaningfully acquired.}
\label{tab:lr-sweep-transition}
\end{table}

At $10^{-6}$, performance changes from T2=79.41/T3=0 at step 30 to
T2=0/T3=18.98 five updates later. At $5\times10^{-7}$, step 60 retains
T2=67.89 while T3 Graph EM is only 2.11\%. Lower learning rates delay the
switch but do not produce a checkpoint with both substantial T2 retention and a
learned T3 policy.

%% file: references.bib
@article{kirkpatrick2017overcoming,
  title={Overcoming catastrophic forgetting in neural networks},
  author={Kirkpatrick, James and Pascanu, Razvan and Rabinowitz, Neil and Veness, Joel and Desjardins, Guillaume and Rusu, Andrei A. and Milan, Kieran and Quan, John and Ramalho, Tiago and Grabska-Barwinska, Agnieszka and others},
  journal={Proceedings of the National Academy of Sciences},
  volume={114},
  number={13},
  pages={3521--3526},
  year={2017}
}

@article{li2017learning,
  title={Learning without forgetting},
  author={Li, Zhizhong and Hoiem, Derek},
  journal={IEEE Transactions on Pattern Analysis and Machine Intelligence},
  volume={40},
  number={12},
  pages={2935--2947},
  year={2018}
}

@inproceedings{lopezpaz2017gradient,
  title={Gradient episodic memory for continual learning},
  author={Lopez-Paz, David and Ranzato, Marc'Aurelio},
  booktitle={Advances in Neural Information Processing Systems},
  volume={30},
  year={2017}
}

@inproceedings{hu2022lora,
  title={LoRA: Low-rank adaptation of large language models},
  author={Hu, Edward J. and Shen, Yelong and Wallis, Phillip and Allen-Zhu, Zeyuan and Li, Yuanzhi and Wang, Shean and Wang, Lu and Chen, Weizhu},
  booktitle={International Conference on Learning Representations},
  year={2022}
}

@inproceedings{shen2024taskbench,
  title={TaskBench: Benchmarking Large Language Models for Task Automation},
  author={Shen, Yongliang and Song, Kaitao and Tan, Xu and Zhang, Wenqi and Ren, Kan and Yuan, Siyu and Lu, Weiming and Li, Dongsheng and Zhuang, Yueting},
  booktitle={Advances in Neural Information Processing Systems},
  volume={37},
  year={2024}
}

@inproceedings{rebuffi2017learning,
  title={Learning multiple visual domains with residual adapters},
  author={Rebuffi, Sylvestre-Alvise and Bilen, Hakan and Vedaldi, Andrea},
  booktitle={Advances in Neural Information Processing Systems},
  volume={30},
  year={2017}
}

@inproceedings{madotto2021continual,
  title={Continual Learning in Task-Oriented Dialogue Systems},
  author={Madotto, Andrea and Lin, Zhaojiang and Zhou, Zhenpeng and Moon, Seungwhan and Crook, Paul and Liu, Bing and Yu, Zhou and Cho, Eunjoon and Fung, Pascale and Wang, Zhiguang},
  booktitle={Proceedings of the 2021 Conference on Empirical Methods in Natural Language Processing},
  pages={7452--7467},
  year={2021},
  doi={10.18653/v1/2021.emnlp-main.590}
}

@inproceedings{wang2024mocl,
  title={Rehearsal-Free Modular and Compositional Continual Learning for Language Models},
  author={Wang, Mingyang and Adel, Heike and Lange, Lukas and Str{\"o}tgen, Jannik and Schuetze, Hinrich},
  booktitle={Proceedings of the 2024 Conference of the North American Chapter of the Association for Computational Linguistics: Human Language Technologies (Volume 2: Short Papers)},
  pages={469--480},
  year={2024},
  publisher={Association for Computational Linguistics},
  doi={10.18653/v1/2024.naacl-short.39},
  url={https://aclanthology.org/2024.naacl-short.39/}
}

@article{zhang2026skilltolora,
  title={Skill-to-LoRA: From Using Skills to Learning Behaviors for Token-Efficient LLM Agents},
  author={Zhang, Tianyi and Qi, Zhonghao},
  journal={arXiv preprint arXiv:2606.16769},
  year={2026}
}

@article{yu2026latentskill,
  title={LatentSkill: From In-Context Textual Skills to In-Weight Latent Skills for LLM Agents},
  author={Yu, Aofan and Zhou, Chenyu and Xu, Tianyi and Guo, Zihan and Shan, Rong and Fu, Zhihui and Wang, Jun and Liu, Weiwen and Yu, Yong and Zhang, Weinan and Lin, Jianghao},
  journal={arXiv preprint arXiv:2606.06087},
  year={2026}
}

@article{guo2026peam,
  title={PEAM: Parametric Embodied Agent Memory through Contrastive Internalization of Experience in Minecraft},
  author={Guo, Yuchen and Gong, Junli and Wang, Weicheng and Cai, Hongmin and Cheung, Yiu-ming and Su, Weifeng},
  journal={arXiv preprint arXiv:2605.27762},
  year={2026}
}

@article{yang2026survey,
  title={A Survey of Agent Skills: Toward Procedural Infrastructure for LLM Agents},
  author={Yang, Cehao and Wu, Xiaojun and Liu, Honghao and Lin, Xueyuan and Xu, Chengjin and Jiang, Xuhui and Sun, Yuanliang and Zhang, Wenjie and Shi, Zhichao and Xu, Yijie and Li, Jia and Xiong, Hui and Guo, Jian},
  journal={Preprints},
  year={2026},
  doi={10.20944/preprints202605.1276.v1}
}

@article{li2026skillsbench,
  title={SkillsBench: Benchmarking How Well Agent Skills Work Across Diverse Tasks},
  author={Li, Xiangyi and others},
  journal={arXiv preprint arXiv:2602.12670},
  year={2026}
}

@article{jiang2026demystifying,
  title={Demystifying Agent Skills: Why They Work---Until They Don't},
  author={Jiang, Zhiyuan and Huang, Fangrui and Xing, Hanwen and Wu, Xander and Gao, Yipeng and Cao, Rui and Wang, Mengdi and Liu, Shilong and Li, Yijiang},
  journal={arXiv preprint arXiv:2608.14036},
  year={2026}
}

@article{lu2026skill0,
  title={SKILL0: In-Context Agentic Reinforcement Learning for Skill Internalization},
  author={Lu, Zhengxi and Yao, Zhiyuan and Wu, Jinyang and Han, Chengcheng and Gu, Qi and Cai, Xunliang and Lu, Weiming and Xiao, Jun and Zhuang, Yueting and Shen, Yongliang},
  journal={arXiv preprint arXiv:2604.02268},
  year={2026}
}

@article{lin2026skillc,
  title={SKILLC: Learning Autonomous Skill Internalization in LLM Agents via Contrastive Credit Assignment},
  author={Lin, Hongxiang and Kuai, Zhirui and Xue, Erpeng and Wang, Lei},
  journal={arXiv preprint arXiv:2605.27899},
  year={2026}
}

@article{zhao2026parametricskills,
  title={Parametric Skills},
  author={Zhao, Xuan and He, Haonan and Yang, Qingyu and Li, Minglei and Ye, Jingqi and Tan, Zelin and Wan, Bo and Ye, Peng},
  journal={arXiv preprint arXiv:2606.30015},
  year={2026}
}

@article{mao2026lifeskill,
  title={Learning While Acting: A Skill-Enhanced Test-Time Co-Evolution Framework for Online Lifelong Learning Agents},
  author={Mao, Bo and Zhou, Jie and Yang, Yutao and Li, Xin and Wei, Xian and Chen, Qin and Wu, Xingjiao and He, Liang},
  journal={arXiv preprint arXiv:2606.04815},
  year={2026}
}

@inproceedings{yao2023react,
  title={ReAct: Synergizing Reasoning and Acting in Language Models},
  author={Yao, Shunyu and Zhao, Jeffrey and Yu, Dian and Du, Nan and Shafran, Izhak and Narasimhan, Karthik and Cao, Yuan},
  booktitle={International Conference on Learning Representations},
  year={2023}
}

@inproceedings{schick2023toolformer,
  title={Toolformer: Language Models Can Teach Themselves to Use Tools},
  author={Schick, Timo and Dwivedi-Yu, Jane and Dess{\`i}, Roberto and Raileanu, Roberta and Lomeli, Maria and Hambro, Eric and Zettlemoyer, Luke and Cancedda, Nicola and Scialom, Thomas},
  booktitle={Advances in Neural Information Processing Systems},
  volume={36},
  year={2023}
}

@article{wang2023voyager,
  title={Voyager: An Open-Ended Embodied Agent with Large Language Models},
  author={Wang, Guanzhi and Xie, Yuqi and Jiang, Yunfan and Mandlekar, Ajay and Xiao, Chaowei and Zhu, Yuke and Fan, Linxi and Anandkumar, Anima},
  journal={arXiv preprint arXiv:2305.16291},
  year={2023}
}

@article{parisi2019continual,
  title={Continual Lifelong Learning with Neural Networks: A Review},
  author={Parisi, German I. and Kemker, Ronald and Part, Jose L. and Kanan, Christopher and Wermter, Stefan},
  journal={Neural Networks},
  volume={113},
  pages={54--71},
  year={2019},
  doi={10.1016/j.neunet.2019.01.012}
}

@article{delange2022continual,
  title={A Continual Learning Survey: Defying Forgetting in Classification Tasks},
  author={De Lange, Matthias and Aljundi, Rahaf and Masana, Marc and Parisot, Sarah and Jia, Xu and Leonardis, Ales and Slabaugh, Gregory and Tuytelaars, Tinne},
  journal={IEEE Transactions on Pattern Analysis and Machine Intelligence},
  volume={44},
  number={7},
  pages={3366--3385},
  year={2022},
  doi={10.1109/TPAMI.2021.3057446}
}

@article{vandeven2019scenarios,
  title={Three Scenarios for Continual Learning},
  author={van de Ven, Gido M. and Tolias, Andreas S.},
  journal={arXiv preprint arXiv:1904.07734},
  year={2019}
}

@article{rusu2016progressive,
  title={Progressive Neural Networks},
  author={Rusu, Andrei A. and Rabinowitz, Neil C. and Desjardins, Guillaume and Soyer, Hubert and Kirkpatrick, James and Kavukcuoglu, Koray and Pascanu, Razvan and Hadsell, Raia},
  journal={arXiv preprint arXiv:1606.04671},
  year={2016}
}

@inproceedings{aljundi2017expertgate,
  title={Expert Gate: Lifelong Learning with a Network of Experts},
  author={Aljundi, Rahaf and Chakravarty, Punarjay and Tuytelaars, Tinne},
  booktitle={Proceedings of the IEEE Conference on Computer Vision and Pattern Recognition},
  pages={3366--3375},
  year={2017}
}

@inproceedings{houlsby2019adapters,
  title={Parameter-Efficient Transfer Learning for {NLP}},
  author={Houlsby, Neil and Giurgiu, Andrei and Jastrzebski, Stanislaw and Morrone, Bruna and de Laroussilhe, Quentin and Gesmundo, Andrea and Attariyan, Mona and Gelly, Sylvain},
  booktitle={Proceedings of the 36th International Conference on Machine Learning},
  pages={2790--2799},
  year={2019}
}

@inproceedings{rebuffi2017icarl,
  title={{iCaRL}: Incremental Classifier and Representation Learning},
  author={Rebuffi, Sylvestre-Alvise and Kolesnikov, Alexander and Sperl, Georg and Lampert, Christoph H.},
  booktitle={Proceedings of the IEEE Conference on Computer Vision and Pattern Recognition},
  pages={2001--2010},
  year={2017}
}

@inproceedings{wang2023orthogonal,
  title={Orthogonal Subspace Learning for Language Model Continual Learning},
  author={Wang, Xiao and Chen, Tianze and Ge, Qiming and Xia, Han and Bao, Rong and Zheng, Rui and Zhang, Qi and Gui, Tao and Huang, Xuanjing},
  booktitle={Findings of the Association for Computational Linguistics: EMNLP 2023},
  pages={10658--10671},
  year={2023},
  doi={10.18653/v1/2023.findings-emnlp.715}
}

@article{shao2024deepseekmath,
  title={DeepSeekMath: Pushing the Limits of Mathematical Reasoning in Open Language Models},
  author={Shao, Zhihong and Wang, Peiyi and Zhu, Qihao and Xu, Runxin and Song, Junxiao and Bi, Xiao and Zhang, Haowei and Zhang, Mingchuan and Li, Y. K. and Wu, Y. and Guo, Daya},
  journal={arXiv preprint arXiv:2402.03300},
  year={2024}
}

@inproceedings{wang2023epi,
  title={Rehearsal-free Continual Language Learning via Efficient Parameter Isolation},
  author={Wang, Zhicheng and Liu, Yufang and Ji, Tao and Wang, Xiaoling and Wu, Yuanbin and Jiang, Congcong and Chao, Ye and Han, Zhencong and Wang, Ling and Shao, Xu and Zeng, Wenqiu},
  booktitle={Proceedings of the 61st Annual Meeting of the Association for Computational Linguistics (Volume 1: Long Papers)},
  pages={10933--10946},
  year={2023},
  publisher={Association for Computational Linguistics},
  doi={10.18653/v1/2023.acl-long.612}
}

@article{yang2025qwen3,
  title={Qwen3 Technical Report},
  author={Yang, An and Li, Anfeng and Yang, Baosong and Zhang, Beichen and Hui, Binyuan and Zheng, Bo and Yu, Bowen and Gao, Chang and Huang, Chengen and Lv, Chenxu and others},
  journal={arXiv preprint arXiv:2505.09388},
  year={2025}
}

@inproceedings{shi2023distracted,
  title={Large Language Models Can Be Easily Distracted by Irrelevant Context},
  author={Shi, Freda and Chen, Xinyun and Misra, Kanishka and Scales, Nathan and Dohan, David and Chi, Ed H. and Sch{\"a}rli, Nathanael and Zhou, Denny},
  booktitle={Proceedings of the 40th International Conference on Machine Learning},
  series={Proceedings of Machine Learning Research},
  volume={202},
  pages={31210--31227},
  year={2023},
  publisher={PMLR}
}

@article{liu2024lostmiddle,
  title={Lost in the Middle: How Language Models Use Long Contexts},
  author={Liu, Nelson F. and Lin, Kevin and Hewitt, John and Paranjape, Ashwin and Bevilacqua, Michele and Petroni, Fabio and Liang, Percy},
  journal={Transactions of the Association for Computational Linguistics},
  volume={12},
  pages={157--173},
  year={2024},
  doi={10.1162/tacl_a_00638}
}
